\documentclass{article}

\usepackage[numbers]{natbib}
\usepackage[preprint]{neurips_2023}
\usepackage[dvipsnames]{xcolor}         
\usepackage[utf8]{inputenc} 
\usepackage[T1]{fontenc}    
\usepackage[colorlinks=true,linkcolor=linkColor,citecolor=linkColor,filecolor=linkColor,urlcolor=linkColor]{hyperref}       
\usepackage{url}            
\usepackage{booktabs}       
\usepackage{amsfonts}       
\usepackage{nicefrac}       
\usepackage{microtype}      
\usepackage{graphicx}
\usepackage{algorithm}

\usepackage{algpseudocode} 
\usepackage{ragged2e}
\usepackage{amsmath}
\usepackage{array}          
\usepackage{tabularx}
\usepackage{colortbl}
\usepackage{float}
\usepackage{comment}
\usepackage{multirow}
\usepackage{diagbox}
\usepackage{makecell}

\usepackage{color}
\usepackage{colortbl}
\usepackage[most]{tcolorbox}
\usepackage{wrapfig}
\usepackage{soul}
\usepackage{listings}
\usepackage{xcolor}

\definecolor{linkColor}{rgb}{0.18,0.39,0.62}
\definecolor{olivegreen}{rgb}{0.0, 0.5, 0.0}
\definecolor{navyblue}{rgb}{0.0, 0.0, 0.5}

\tcbset {
  base/.style={
    arc=0mm, 
    bottomtitle=-0.25mm,
    boxrule=0mm,
    colbacktitle=black!10!white, 
    coltitle=black, 
    fonttitle=\bfseries, 
    left=2.5mm,
    leftrule=1mm,
    right=3.5mm,
    title={#1},
    toptitle=0.25mm,
  }
}
\usepackage{listings}
\usepackage{xcolor}

\newcommand\bitnet{\text{BitNet b1.58}}

\usepackage{multirow}
\usepackage{amsmath}
\usepackage{capt-of}
\usepackage{tabularx}
\usepackage{epsfig}
\usepackage{amssymb}
\usepackage{amsfonts}
\usepackage{booktabs}
\usepackage{scalerel}
\usepackage[inline]{enumitem}
\usepackage{listings}
\usepackage{varwidth}
\usepackage[export]{adjustbox}
\usepackage{tikz}
\usetikzlibrary{tikzmark}

\usepackage{stmaryrd}
\usepackage{bbm}
\usepackage{wrapfig}
\usepackage{pifont}
\usepackage[noabbrev]{cleveref}

\definecolor{deepblue}{rgb}{0,0,0.5}
\definecolor{officeblue}{RGB}{0,102,204}
\definecolor{deepred}{rgb}{0.6,0,0}
\definecolor{deepgreen}{rgb}{0,0.5,0}
\definecolor{mybrickred}{RGB}{182,50,28}

\definecolor{fillcolor}{RGB}{216,217,252}

\usepackage{etoolbox}
\usepackage{framed}

\newif\ifxetexorluatex
\ifxetex
  \xetexorluatextrue
\else
  \ifluatex
    \xetexorluatextrue
  \else
    \xetexorluatexfalse
  \fi
\fi
\newcommand*\quotesize{60} 
\newcommand*{\openquote}
   {\tikz[remember picture,overlay,xshift=-4ex,yshift=-2.5ex]
   \node (OQ) {\fontsize{\quotesize}{\quotesize}\selectfont``};\kern0pt}

\newcommand*{\closequote}[1]
  {\tikz[remember picture,overlay,xshift=4ex,yshift={#1}]
   \node (CQ) {\fontsize{\quotesize}{\quotesize}\selectfont''};}

\colorlet{shadecolor}{white}

\newcommand*\shadedauthorformat{\emph} 

\newcommand*\authoralign[1]{%
  \if#1l
    \def\authorfill{}\def\quotefill{\hfill}
  \else
    \if#1r
      \def\authorfill{\hfill}\def\quotefill{}
    \else
      \if#1c
        \gdef\authorfill{\hfill}\def\quotefill{\hfill}
      \else\typeout{Invalid option}
      \fi
    \fi
  \fi}
{\authoralign{#1}
\ifblank{#2}
   {\def\shadequoteauthor{}\def\yshift{-2ex}\def\quotefill{\hfill}}
   {\def\shadequoteauthor{\par\authorfill\shadedauthorformat{#2}}\def\yshift{2ex}}
\begin{snugshade}\begin{quote}\openquote}
{\shadequoteauthor\quotefill\closequote{\yshift}\end{quote}\end{snugshade}}

\newfloat{Code}{htbp}{Code}

\usepackage{amsmath,amsfonts,bm}

\def\eqref#1{equation~(\ref{#1})}

\def\1{\bm{1}}

\DeclareMathAlphabet{\mathsfit}{\encodingdefault}{\sfdefault}{m}{sl}
\SetMathAlphabet{\mathsfit}{bold}{\encodingdefault}{\sfdefault}{bx}{n}

\title{1-bit AI Infra: Part 1.1, Fast and \colorbox{gray!30}{Lossless} \\ BitNet b1.58 Inference on CPUs }

\vspace{-2.8cm}
\author{Microsoft Research \\   \href{https://aka.ms/GeneralAI}{https://aka.ms/GeneralAI}}

\vspace{-2.8cm}
\author{
Jinheng Wang\thanks{~Equal contribution. $\diamond$ Corresponding author. T. Song, S. Mao, S. Ma, Y. Xia, F. Wei are with Microsoft Research. J. Wang and H. Zhou are with Peking University. H. Wang is with University of Chinese Academy of Sciences.},~~~~Hansong Zhou\footnotemark[1],~~~~Ting Song\footnotemark[1],~~~~Shaoguang Mao,\\~~~~\textbf{Shuming Ma,}~~~~\textbf{Hongyu Wang,}~~~~\textbf{Yan Xia,}~~~~\textbf{Furu Wei}$^{\diamond}$
\\
Microsoft Research \\ 
{\href{https://aka.ms/GeneralAI}{https://aka.ms/GeneralAI}}
\\}

\begin{document}

\maketitle
\vspace{-0.5cm}
\begin{abstract}
\vspace{-0.25cm}
Recent advances in 1-bit Large Language Models (LLMs), such as BitNet~\cite{wang2023bitnet} and BitNet b1.58~\cite{ma2024era}, present a promising approach to enhancing the efficiency of LLMs in terms of speed and energy consumption. These developments also enable \textbf{local LLM} deployment across a broad range of devices. In this work, we introduce \textbf{\colorbox{gray!30}{bitnet.cpp}}, a tailored software stack designed to unlock the full potential of 1-bit LLMs. Specifically, we develop a set of kernels to support \textbf{fast} and \textbf{lossless} inference of ternary BitNet b1.58 LLMs on CPUs. Extensive experiments demonstrate that bitnet.cpp achieves significant speedups, ranging from 2.37x to 6.17x on x86 CPUs and from 1.37x to 5.07x on ARM CPUs, across various model sizes. The code is available at \href{https://github.com/microsoft/BitNet}{aka.ms/bitnet}.

\end{abstract}
\begin{figure}[ht]
\vspace{-0.7cm}
    \centering
\includegraphics[width=\linewidth]{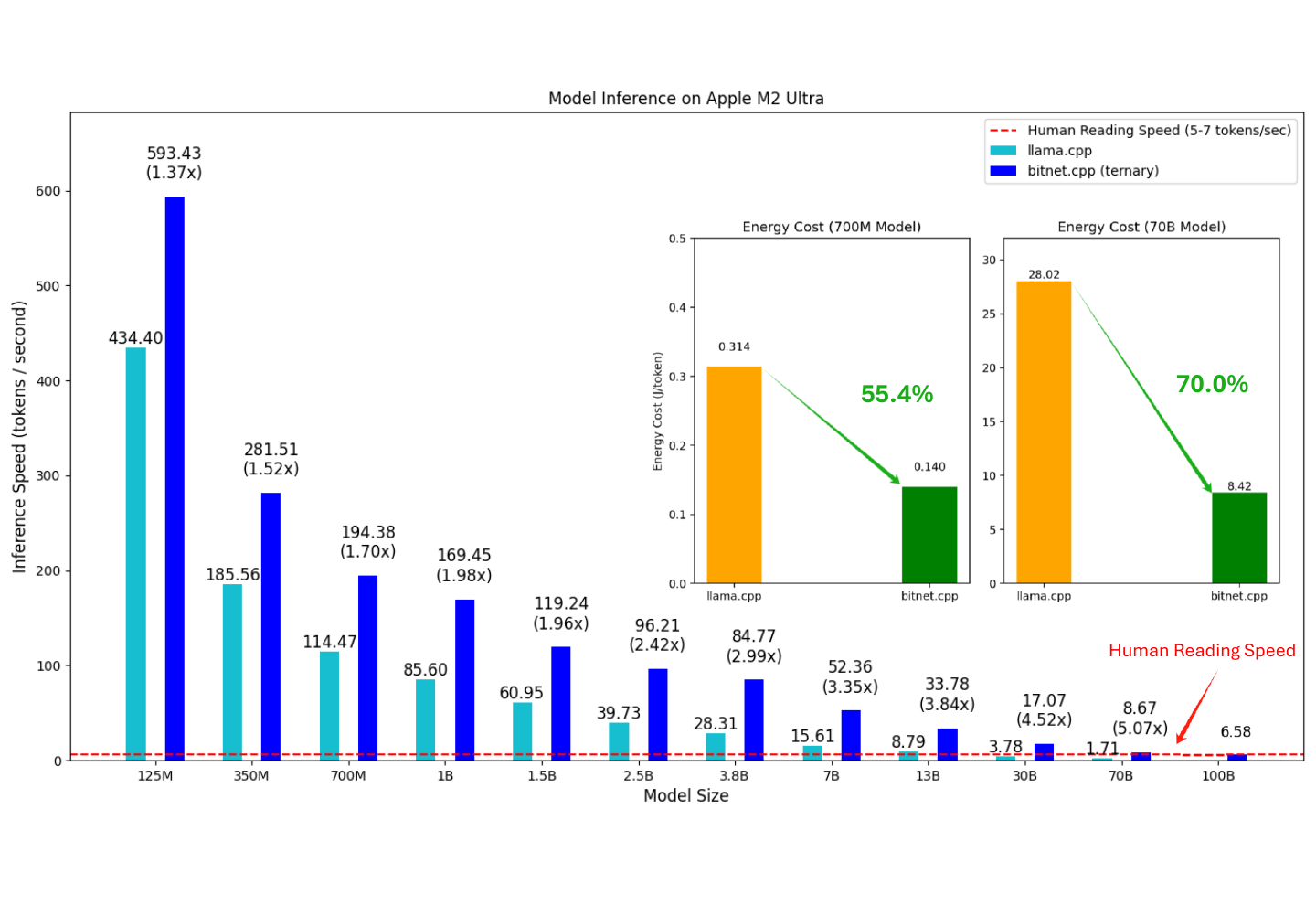}
    \vspace{-1.4cm}
\caption{Comparison of \textbf{inference speed} and \textbf{energy consumption} for various \bitnet{} model sizes on an Apple M2 Ultra (ARM CPU) using llama.cpp(fp16) \cite{llamacpp} versus bitnet.cpp (ternary kernels). The results demonstrate that bitnet.cpp can achieve human reading speed, even for a 100B model on a single CPU. Notably, bitnet.cpp significantly reduces energy consumption across different model sizes.} 
    \label{fig:performance}
\end{figure}

\newpage
\section{bitnet.cpp}

bitnet.cpp is an inference framework for 1-bit LLMs (e.g., \bitnet{} models). It provides \textbf{lossless inference} while optimizing both \textbf{speed and energy consumption}. The initial release of bitnet.cpp supports inference on CPUs.

As illustrated in Figure \ref{fig:performance}, bitnet.cpp achieves speedups ranging from 1.37x to 5.07x on ARM CPUs, with larger models experiencing greater performance 
gains. Additionally, it reduces energy consumption by 55.4\% to 70.0\%, further boosting overall efficiency. On x86 CPUs, speedups range from 2.37x to 6.17x with energy reductions between 71.9\% and 82.2\%. Furthermore, bitnet.cpp can run a 100B \bitnet{} model on a single CPU, achieving speeds comparable to human reading (5-7 tokens per second)\cite{brysbaert2019many},  thus significantly enhancing the potential for running LLMs on local devices.

To start using bitnet.cpp for inference, follow these steps:

\begin{figure}[htbp]
\centering
\begin{minipage}[t]{0.85\columnwidth}
\begin{lstlisting}[language=bash]
  # Clone the repo
  git clone git clone --recursive https://github.com/microsoft/BitNet.git
  cd BitNet

  # Create a new conda environment (Recommended)
  conda create -n bitnet-cpp python=3.9
  conda activate bitnet-cpp
  pip install -r requirements.txt

  # Download the model from Hugging Face, convert it to quantized gguf format, and build the project
  # These models were neither trained nor released by Microsoft. We used them to demonstrate the inference capabilities of bitnet.cpp
  python setup_env.py --hf-repo HF1BitLLM/Llama3-8B-1.58-100B-tokens -q i2_s
  
  # Or you can manually download the model and run it using a local path
  huggingface-cli download HF1BitLLM/Llama3-8B-1.58-100B-tokens --local-dir models/Llama3-8B-1.58-100B-tokens
  python setup_env.py -md models/Llama3-8B-1.58-100B-tokens -q i2_s

  # Run inference with the quantized model, use -m to specify the model path, -p to specify the prompt
  python run_inference.py -m models/Llama3-8B-1.58-100B-tokens/ggml-model-i2_s.gguf -p "Daniel went back to the the the garden. Mary travelled to the kitchen. Sandra journeyed to the kitchen. Sandra went to the hallway. John went to the bedroom. Mary went back to the garden. Where is Mary?\nAnswer:"
\end{lstlisting}
\end{minipage}
\end{figure}

\section{Optimized Kernels for 1.58-bit Models}

bitnet.cpp offers a suite of optimized kernels, including I2\_S, TL1 and TL2. The kernels are designed for fast and lossless inference of 1.58-bit models on both x86 and ARM architectures. 

\begin{table}[H]
\centering
\begin{tabular}{|c|c|}
\hline
\textbf{Unpack} & \textbf{Pack} \\ \hline
-1  & 00 \\ \hline
0  & 01 \\ \hline
1  & 10 \\ \hline
\end{tabular}
\vspace{0.3cm}
\caption{I2\_S Kernel transforms each full-precision weight into a 2-bit value to save memory and bandwidth. When performing computation, the 2-bit weights are unpacked to the original values.}
\label{tab:i2_s}
\end{table}

\textbf{I2\_S Kernel} adopts the vanilla multiply-then-addition manner to perform the matrix multiplication. As shown in Table~\ref{tab:i2_s}, it transforms each full-precision weight into a 2-bit representation offline. During computation, it transforms the weights back to their original values and performs the vanilla GEMV operations.
We recommend using it with sufficient threads, since it allows the compiler to generate efficient pipelined instruction sequences.
 
\begin{table}[H]
\centering
\begin{tabular}{|>{\centering\arraybackslash}p{1cm}|>{\centering\arraybackslash}p{1cm}|>{\centering\arraybackslash}p{1.5cm}|}
\hline
\multicolumn{2}{|c|}{\textbf{Unpack}} & \textbf{Pack} \\ \hline
-1 & -1 & 0000 \\ \hline
-1 & 0 & 0001 \\ \hline
-1 & 1 & 0010 \\ \hline
0 & -1 & 0011 \\ \hline
0 & 0 & 0100 \\ \hline
0 & 1 & 0101 \\ \hline
1 & -1 & 0110 \\ \hline
1 & 0 & 0111 \\ \hline
1 & 1 & 1000 \\ \hline
\end{tabular}
\vspace{0.3cm}
    \caption{TL1 Kernel transforms every two full-precision weights into 4-bit index and performs LUT computation.}
    \label{tab:tl1}
\end{table}

\textbf{TL1 Kernel} preprocesses every two full-precision weights by packing them into 4-bit index (see Table~\ref{tab:tl1}), and pre-computes their corresponding activations into $3^2 = 9$ values. The index-value pairs are stored in a lookup table to perform LUT computation~\cite{park2022lut, wei2024t}. GEMV processing is performed using an int16 LUT and accumulation through addition. We recommend using it with a limited number of threads when serving large models.

\begin{table}[H]
\centering
\begin{tabular}{|>{\centering\arraybackslash}p{1cm}|>{\centering\arraybackslash}p{1cm}|>{\centering\arraybackslash}p{1cm}|>{\centering\arraybackslash}p{1.5cm}|}
\hline
\multicolumn{3}{|c|}{\textbf{Unpack}} & \textbf{Pack} \\ \hline
-1 & -1 & -1 & \textcolor{red}{1}\space1101 \\ \hline
-1 & -1 & 0 & \textcolor{red}{1}\space1100 \\ \hline
-1 & -1 & 1 & \textcolor{red}{1}\space1011 \\ \hline
-1 & 0 & -1 & \textcolor{red}{1}\space1010 \\ \hline
\multicolumn{4}{|c|}{...}\\
\hline
0 & 0 & 0 & \textcolor{red}{0}\space0000
 \\ \hline
\multicolumn{4}{|c|}{...}
\\ \hline
1 & 0 & 1 &  \textcolor{red}{0}\space1010\\ \hline
1 & 1 & -1 & \textcolor{red}{0}\space1011 \\ \hline
1 & 1 & 0 & \textcolor{red}{0}\space1100 \\ \hline
1 & 1 & 1 & \textcolor{red}{0}\space1101 \\ \hline
\end{tabular}
\vspace{0.3cm}
    \caption{TL2 Kernel compresses every three full-precision weights into a 1-bit sign (\textcolor{red}{0} or  \textcolor{red}{1}) and a 4-bit index. }
    \label{tab:tl2}
\end{table}
\textbf{TL2 Kernel} is similar to TL1. The major difference is that it compresses every three weights into a 5-bit index, while TL1 compresses every two weights into a 4-bit index. Therefore, TL2 achieves a higher compression ratio than TL1. We recommend using it in environments with limited memory or bandwidth, since it employs LUT and reduces model size by 1/6 compared to TL1 Kernel, thereby lowering bandwidth requirements. 

\section{Evaluation}
\subsection{Inference Performance}
We evaluated bitnet.cpp in terms of both \textbf{inference speed} and \textbf{energy cost}. Comprehensive tests were conducted on models with various parameter sizes, ranging from 125M to 100B. specific configurations for each model are detailed in the Appendix \ref{app:model_config}. These sizes represent popular LLM configurations. Additionally, systematic tests were performed on both ARM and x86 architectures. For ARM, we used a Mac Studio with an Apple M2 Ultra processor and 64GB of memory for end-to-end tests. For x86, a Surface Laptop Studio 2 with an Intel Core i7-13700H processor (14 cores, 20 threads) and 64GB of memory was used. 

We tested two scenarios for each device: one with inference limited to two threads, and the other without thread restrictions, reporting the optimal inference speed. It was to consider the limited thread availability on local devices, providing a more accurate performance assessment of \bitnet{} in local environments.

\textbf{Inference Speed:} Table~\ref{tab:speed_thread=unlimited} and Table~\ref{tab:speed_thread=2} demonstrate significant performance advantages of bitnet.cpp over llama.cpp on both ARM (Apple M2) and x86 (Intel i7-13700H) architectures, especially as model sizes increase. bitnet.cpp consistently outpaces llama.cpp, with speedups ranging from 1.37x to 6.46x, depending on the model and architecture. On the Apple M2, speedups peak at 5.07x in the unlimited thread scenario, while on the Intel i7-13700H, bitnet.cpp achieves up to 6.46x in thread-limited scenarios, making it particularly effective for local inference on resource-constrained systems.

The performance gap widens as models scale up, with larger models (13B and above) benefiting the most from bitnet.cpp's optimizations. On the Intel i7-13700H, bitnet.cpp provides substantial speed improvements, making it well-suited for x86 architecture, even with limited threads. While smaller models (125M to 1B) also see meaningful gains, the advantages of bitnet.cpp become especially critical for larger, more complex models, underscoring its efficiency in handling demanding workloads. Adding to the observed performance differences, bandwidth limitations play a significant role in the varying efficacy of bitnet.cpp across different architectures, particularly when comparing the Apple M2 and Intel i7-13700H. Due to the larger bandwidth of the Apple M2, it achieves significantly faster speed improvements with bitnet.cpp compared to the Intel i7-13700H, especially when running larger models.

\begin{table}[ht]
    \centering
    \resizebox{\textwidth}{!}{
    \renewcommand{\arraystretch}{1.2}
    \begin{tabular}{l|l|ccccccccccccc}
        \toprule
        CPU & Kernel & 125M & 350M & 700M & 1B & 1.5B & 2.5B & 3.8B & 7B & 13B & 30B & 70B & 100B \\
        \midrule
        \multirow{3}{*}{APPLE M2} 
        & llama.cpp & 434.40 & 186.56 & 114.47 & 85.60 & 60.95 & 39.73 & 28.31 & 15.61 & 8.79 & 3.78 & 1.71 & N/A \\
        \cline{2-14}
        & \multirow{2}{*}{bitnet.cpp} & \textbf{593.43} & \textbf{281.51} & \textbf{194.38} & \textbf{169.45} & \textbf{119.24} & \textbf{96.21} & \textbf{84.77} & \textbf{52.36} & \textbf{33.78} & \textbf{17.07} & \textbf{8.67} & \textbf{6.58} \\
        &  & 
        (1.37x) & (1.51x) & (1.70x) & (1.98x) & (1.96x) & (2.42x) & (2.99x) & (3.35x) & (3.84x) & (4.51x) & (5.07x) & (N/A) \\
        \midrule
        \multirow{3}{*}{\makecell[l]{Intel i7-13700H \\ 20C 64G}} 
        & llama.cpp & 164.04 & 56.67 & 30.73 & 22.31 & 15.02 & 11.07 & 5.85 & 3.30 & 1.78 & N/A & N/A & N/A \\
        \cline{2-14}
        & \multirow{2}{*}{bitnet.cpp} & \textbf{389.08} & \textbf{172.95} & \textbf{119.08} & \textbf{86.50} & \textbf{67.12} & \textbf{46.33} & \textbf{30.51} & \textbf{18.75} & \textbf{10.99} & \textbf{5.10} & \textbf{2.44} & \textbf{1.70} \\
        &  & 
        (2.37x) & (3.05x) & (3.88x) & (3.88x) & (4.47x) & (4.19x) & (5.22x) & (5.68x) & (6.17x) & (N/A) & (N/A) & (N/A) \\
        \bottomrule
    \end{tabular}}
    \caption{Comparison of inference speed across different CPUs (Unit: Tokens/Second) in an unlimited thread setting. "N/A" indicates that the tested CPU cannot host the specified model size with the given kernel.}
    \label{tab:speed_thread=unlimited}
\end{table}

\begin{table}[ht]
    \centering
    \resizebox{\textwidth}{!}{
    \renewcommand{\arraystretch}{1.2}
    \begin{tabular}{l|l|ccccccccccccc}
        \toprule
        CPU & Kernel & 125M & 350M & 700M & 1B & 1.5B & 2.5B & 3.8B & 7B & 13B & 30B & 70B & 100B \\
        \midrule
        \multirow{3}{*}{APPLE M2} 
        & llama.cpp & 251.95 & 95.18 & 53.93 & 41.36 & 26.67 & 17.61 & 11.88 & 6.71 & 3.73 & 1.60 & 0.71 & N/A \\
        \cline{2-14}
        & \multirow{2}{*}{bitnet.cpp} & \textbf{401.76} & \textbf{168.88} & \textbf{96.34} & \textbf{79.23} & \textbf{56.31} & \textbf{37.29} & \textbf{26.75} & \textbf{15.26} & \textbf{8.75} & \textbf{3.89} & \textbf{1.76} & \textbf{1.27} \\
        &  & 
        (1.59x) & (1.77x) & (1.79x) & (1.92x) & (2.11x) & (2.12x) & (2.25x) & (2.27x) & (2.35x) & (2.43x) & (2.48x) & (N/A) \\
        \midrule
        \multirow{3}{*}{\makecell[l]{Intel i7-13700H \\ 20C 64G}} 
        & llama.cpp & 119.84 & 41.57 & 18.56 & 13.92 & 8.99 & 6.95 & 3.49 & 1.92 & 1.30 & N/A & N/A & N/A \\
        \cline{2-14}
        & \multirow{2}{*}{bitnet.cpp} & \textbf{316.35} & \textbf{137.68} & \textbf{80.13} & \textbf{57.76} & \textbf{44.69} & \textbf{29.41} & \textbf{20.51} & \textbf{12.41} & \textbf{7.09} & \textbf{3.23} & \textbf{1.51} & \textbf{0.97} \\
        &  & 
        (2.64x) & (3.31x) & (4.32x) & (4.15x) & (4.97x) & (4.23x) & (5.88x) & (6.46x) & (5.45x) & (N/A) & (N/A) & (N/A) \\
        \bottomrule
    \end{tabular}}
    \caption{Comparison of inference speed across different CPUs (Unit: Tokens/Second) in a thread-limited setting, where the number of available inference threads is set to 2. "N/A" indicates that the tested CPU cannot host the specified model size with the given kernel.}
    \label{tab:speed_thread=2}
\end{table}

\textbf{Energy Cost:} We ran 700M, 7B and 70B models and reported the energy cost (J/token) with the best inference speed in the unlimited thread setting. Table \ref{tab:energy_thread_unlimited}  demonstrates a clear advantage of bitnet.cpp in reducing energy consumption. For the Apple M2, bitnet.cpp reduces energy usage by 55.4\% to 70.0\% depending on the model size. As model size increases, bitnet.cpp's energy efficiency becomes more pronounced, with the largest model (70B) showing a 70.0\% reduction in energy consumption compared to llama.cpp. This highlights bitnet.cpp's ability to deploy large-scale inference more efficiently, both in terms of speed and energy usage, which is crucial for energy-constrained environments such as mobile devices or edge computing.

On the Intel i7-13700H, energy savings with bitnet.cpp are even more dramatic, ranging from 71.9\% to 82.2\% for models up to 7B. Although energy consumption data for the 70B model on the Intel CPU is unavailable, the results for smaller models clearly show that bitnet.cpp can significantly lower the energy demands of large language model inference on high-performance, multi-core processors.

\begin{table}[H]
    \centering
    \resizebox{0.5\textwidth}{!}{
    \renewcommand{\arraystretch}{1.2}
    \begin{tabular}{llccc}
        \toprule
        CPU & Kernel & 700M & 7B & 70B \\
        \midrule
        \multirow{3}{*}{APPLE M2} 
        & llama.cpp & 0.314 & 3.013 & 28.02 \\
        & bitnet.cpp & \textbf{0.140} & \textbf{1.068} & \textbf{8.42} \\\cline{2-5}
        & saving & 55.4\% & 64.6\% & 70.0\% \\
        \midrule
        \multirow{3}{*}{\makecell[l]{Intel i7-13700H \\ 20C 64G}} 
        & llama.cpp & 1.367 & 11.305 & N/A \\
        & bitnet.cpp & \textbf{0.384} & \textbf{2.017} & \textbf{17.33} \\\cline{2-5}
        & saving & 71.9\% & 82.2\% & N/A \\
        \bottomrule
    \end{tabular}}
    \vspace{0.2cm}
    \caption{Comparison of Energy Costs Across CPUs (Unit: J/Token). "N/A" indicates that the specific model size cannot be hosted on the tested CPU with the given kernel.}
    \label{tab:energy_thread_unlimited}
\end{table}

\subsection{Inference Accuracy}

The bitnet.cpp framework enables lossless inference for ternary \bitnet{} LLMs. To evaluate \textbf{inference accuracy}, we randomly selected 1,000 prompts from WildChat \cite{zhao2024wildchat} and compared the outputs generated by bitnet.cpp and llama.cpp to those produced by an FP32 kernel. The evaluation was conducted on a token-by-token basis, with a maximum of 100 tokens per model output, considering an inference sample lossless only if it exactly matched the full-precision output. 

This evaluation used a 700M \bitnet{} model\footnote{We used \href{https://huggingface.co/1bitLLM/bitnet_b1_58-large}{bitnet\_b1\_58-large} available on HuggingFace to demonstrate the inference capabilities of bitnet.cpp. This model and the 8B \bitnet{}\cite{mohamed1.58} in the quick start were \textcolor{red}{neither trained nor released by Microsoft}.}.  The results confirm that bitnet.cpp achieves accurate, lossless inference for 1-bit LLMs.
\begin{table}[h]
    \centering \renewcommand{\arraystretch}{1.2}
    \begin{tabular}{c|cc|ccc}
        \toprule
        \multirow{2}{*}{Kernel} &\multicolumn{2}{c|}{llama.cpp} & \multicolumn{3}{c}{bitnet.cpp} \\ \cline{2-6}
         &  TQ1\_0 & TQ2\_0 & l2\_S & TL1 & TL2 \\ \hline
        Accuracy &  1.4\% & 1.4\% & 100\% & 100\% & 100\% \\ 
        \bottomrule
    \end{tabular}
    \vspace{0.2cm}
    \caption{Comparison of inference accuracy between llama.cpp and bitnet.cpp. TQ1\_0 and TQ2\_0 are kernels in llama.cpp, while l2\_S, TL1, and TL2 are kernels in bitnet.cpp. Accuracy indicates the proportion of lossless inference samples, where outputs matched exactly with the full-precision baseline.}

\end{table}

\section{Future Work}
We are expanding bitnet.cpp to support a broader range of platforms and devices, including mobile devices (e.g., iPhone and Android), NPUs, and GPUs. We will also work on 1-bit LLM training optimization in the future. Furthermore, we are interested in the co-design of customized hardware and software stacks for 1-bit LLMs.

\section*{Acknowledgement}
We express our sincere gratitude to Georgi Gerganov and the entire llama.cpp community, whose work served as the foundation for our implementation of BitNet b1.58 kernels. Additionally, we extend our thanks to our colleagues and fellow interns at Microsoft Research Asia for their invaluable discussions and feedback. In particular, we acknowledge Jiangyu Wei, Shijie Cao, and Ting Cao for introducing the LUT method for low-bit LLM inference on CPUs in their T-MAC work. Xiaoyan Hu provided extensive system and device support, as well as insightful discussions on operating system integration.

\bibliographystyle{alpha}
\bibliography{custom}

\newcommand{\etalchar}[1]{$^{#1}$}
\begin{thebibliography}{MSvWW24}

\bibitem[Bry19]{brysbaert2019many}
Marc Brysbaert.
\newblock How many words do we read per minute? a review and meta-analysis of reading rate.
\newblock {\em Journal of memory and language}, 109:104047, 2019.

\bibitem[lla]{llamacpp}
llama.cpp.
\newblock \url{https://github.com/ggerganov/llama.cpp}.

\bibitem[MSvWW24]{mohamed1.58}
Mohamed Mekkouri, Marc Sun, Leandro von Werra, and Thomas Wolf.
\newblock 1.58-bit llm: A new era of extreme quantization, 2024.

\bibitem[MWM{\etalchar{+}}24]{ma2024era}
Shuming Ma, Hongyu Wang, Lingxiao Ma, Lei Wang, Wenhui Wang, Shaohan Huang, Li~Dong, Ruiping Wang, Jilong Xue, and Furu Wei.
\newblock The era of 1-bit llms: All large language models are in 1.58 bits.
\newblock {\em arXiv preprint arXiv:2402.17764}, 2024.

\bibitem[PPK{\etalchar{+}}22]{park2022lut}
Gunho Park, Baeseong Park, Minsub Kim, Sungjae Lee, Jeonghoon Kim, Beomseok Kwon, Se~Jung Kwon, Byeongwook Kim, Youngjoo Lee, and Dongsoo Lee.
\newblock Lut-gemm: Quantized matrix multiplication based on luts for efficient inference in large-scale generative language models.
\newblock {\em arXiv preprint arXiv:2206.09557}, 2022.

\bibitem[WCC{\etalchar{+}}24]{wei2024t}
Jianyu Wei, Shijie Cao, Ting Cao, Lingxiao Ma, Lei Wang, Yanyong Zhang, and Mao Yang.
\newblock T-mac: Cpu renaissance via table lookup for low-bit llm deployment on edge.
\newblock {\em arXiv preprint arXiv:2407.00088}, 2024.

\bibitem[WMD{\etalchar{+}}23]{wang2023bitnet}
Hongyu Wang, Shuming Ma, Li~Dong, Shaohan Huang, Huaijie Wang, Lingxiao Ma, Fan Yang, Ruiping Wang, Yi~Wu, and Furu Wei.
\newblock Bitnet: Scaling 1-bit transformers for large language models.
\newblock {\em arXiv preprint arXiv:2310.11453}, 2023.

\bibitem[ZRH{\etalchar{+}}24]{zhao2024wildchat}
Wenting Zhao, Xiang Ren, Jack Hessel, Claire Cardie, Yejin Choi, and Yuntian Deng.
\newblock Wildchat: 1m chat{GPT} interaction logs in the wild.
\newblock In {\em The Twelfth International Conference on Learning Representations}, 2024.

\end{thebibliography}

\appendix
\section{Model Config}\label{app:model_config}
The tested models are dummy setups used in a research context to demonstrate the inference performance of bitnet.cpp. The specific configuration is as follows:
\begin{lstlisting}
{
    "125M": {
        "hidden_size": 768,
        "intermediate_size": 3072,
        "num_hidden_layers": 11,
        "num_attention_heads": 12
    },
    "350M": {
        "hidden_size": 1024,
        "intermediate_size": 3072,
        "num_hidden_layers": 24,
        "num_attention_heads": 16
    },
    "700M": {
        "hidden_size": 1536,
        "intermediate_size": 4096,
        "num_hidden_layers": 24,
        "num_attention_heads": 16
    },
    "1B": {
        "hidden_size": 2048,
        "intermediate_size": 3584,
        "num_hidden_layers": 24,
        "num_attention_heads": 32
    },
    "1.5B": {
        "hidden_size": 1536,
        "intermediate_size": 9216,
        "num_hidden_layers": 28,
        "num_attention_heads": 32
    },
    "2.5B": {
        "hidden_size": 2560,
        "intermediate_size": 6912,
        "num_hidden_layers": 30,
        "num_attention_heads": 20
    },
    "3.8B": {
        "hidden_size": 3840,
        "intermediate_size": 8192,
        "num_hidden_layers": 24,
        "num_attention_heads": 32
    },
    "7B": {
        "hidden_size": 4096,
        "intermediate_size": 12032,
        "num_hidden_layers": 32,
        "num_attention_heads": 32
    },
    "13B": {
        "hidden_size": 5120,
        "intermediate_size": 13824,
        "num_hidden_layers": 40,
        "num_attention_heads": 40
    },
    "30B": {
        "hidden_size": 6656,
        "intermediate_size": 16384,
        "num_hidden_layers": 60,
        "num_attention_heads": 52
    },
    "70B": {
        "hidden_size": 8192,
        "intermediate_size": 24576,
        "num_hidden_layers": 80,
        "num_attention_heads": 64
    },
    "100B": {
        "hidden_size": 8192,
        "intermediate_size": 45568,
        "num_hidden_layers": 72,
        "num_attention_heads": 64
    }
}
\end{lstlisting}
We hope the release of bitnet.cpp will inspire the development of 1-bit LLMs in large-scale settings in terms of model size and training tokens.

\section{Statement of Contribution}
All co-authors contributed to discussions, provided input on various aspects of the project, and assisted with experimental design, paper writing, and resource coordination. In addition to these contributions, J. Wang and T. Song crafted multiple kernels and laid out the overarching architecture for the BitNet inference framework. J. Wang took on the full task of kernel implementation. H. Zhou invested considerable time running experiments focused on inference speed, accuracy, and energy consumption.

\end{document}